\documentclass[sigconf, screen]{acmart}
\AtBeginDocument{%
  }

\setcopyright{acmlicensed}
\copyrightyear{2026}
\acmYear{2026}
\acmDOI{XXXXXXX.XXXXXXX}
\acmISBN{978-1-4503-XXXX-X/2018/06}

\usepackage{multirow}
\usepackage{tabularx}  
\usepackage{booktabs}  

\usepackage{hyperref}

\begin{document}

\title{DaViNCi: A Dataset Towards Outdoor Vision-and-Language Navigation with Continuous Actions and Dynamic Elements}

\author{Zihao Xie}
\authornote{Both authors contributed equally to this research.}
\email{xzh031202@sjtu.edu.cn}
\affiliation{%
  \institution{Shanghai Jiao Tong University}
  \city{Shanghai}
  \country{China}
}

\author{Pingrui Lai}
\authornotemark[1]
\email{laipingrui@sjtu.edu.cn}
\affiliation{%
  \institution{Shanghai Jiao Tong University}
  \city{Shanghai}
  \country{China}
}

\author{Yitong Wu}
\email{maca8ka@sjtu.edu.cn}
\affiliation{%
  \institution{Shanghai Jiao Tong University}
  \city{Shanghai}
  \country{China}}

\author{Hua Yang}
\email{hyang@sjtu.edu.cn}
\authornote{Corresponding author}
\affiliation{%
  \institution{Shanghai Jiao Tong University}
  \city{Shanghai}
  \country{China}}

\renewcommand{\shortauthors}{Z Xie et al.}

\begin{abstract}
  Vision-and-Language Navigation (VLN) has progressively expanded from indoor to outdoor environments. However, existing outdoor VLN datasets still rely on fixed discrete topological graphs for construction. It fails to align with the rapidly changing real-world outdoor environments and impedes the sim-to-real transfer of VLN agents. To address this limitation, we propose DaViNCi (\textbf{D}yn\textbf{a}mic \textbf{Vi}sion-and-Language \textbf{N}avigation in \textbf{C}ont\textbf{i}nuous Environment), the first outdoor VLN dataset that simultaneously introduces both continuous and dynamic factors. The agent not only moves in the outdoor environment using continuous actions but is also required to handle unpredictable dynamic elements. The dataset encompasses six distinct maps with a total of 6,933 trajectories. Through comprehensive comparative experiments, we find that the success rate on DaViNCi decreased by more than 10\% in discrete environments compared to previous datasets. And there is an even greater decline in continuous settings, demonstrating the challenge of DaViNCi. Furthermore, we clarify the impact of action granularity and dynamic elements. These results demonstrate the practical value of DaViNCi in advancing outdoor VLN toward more realistic environments. The website is \url{https://xzh0312.github.io/DaViNCi/}.
\end{abstract}

\ccsdesc[500]{Computer systems organization~Robotics}

\keywords{Outdoor Vision-and-Language Navigation, Continuous Environment, Dynamic Elements, Reinforcement Learning}

\received{20 February 2007}
\received[revised]{12 March 2009}
\received[accepted]{5 June 2009}

\maketitle

\begin{figure}[t]
  \centering
  \includegraphics[width=\linewidth]{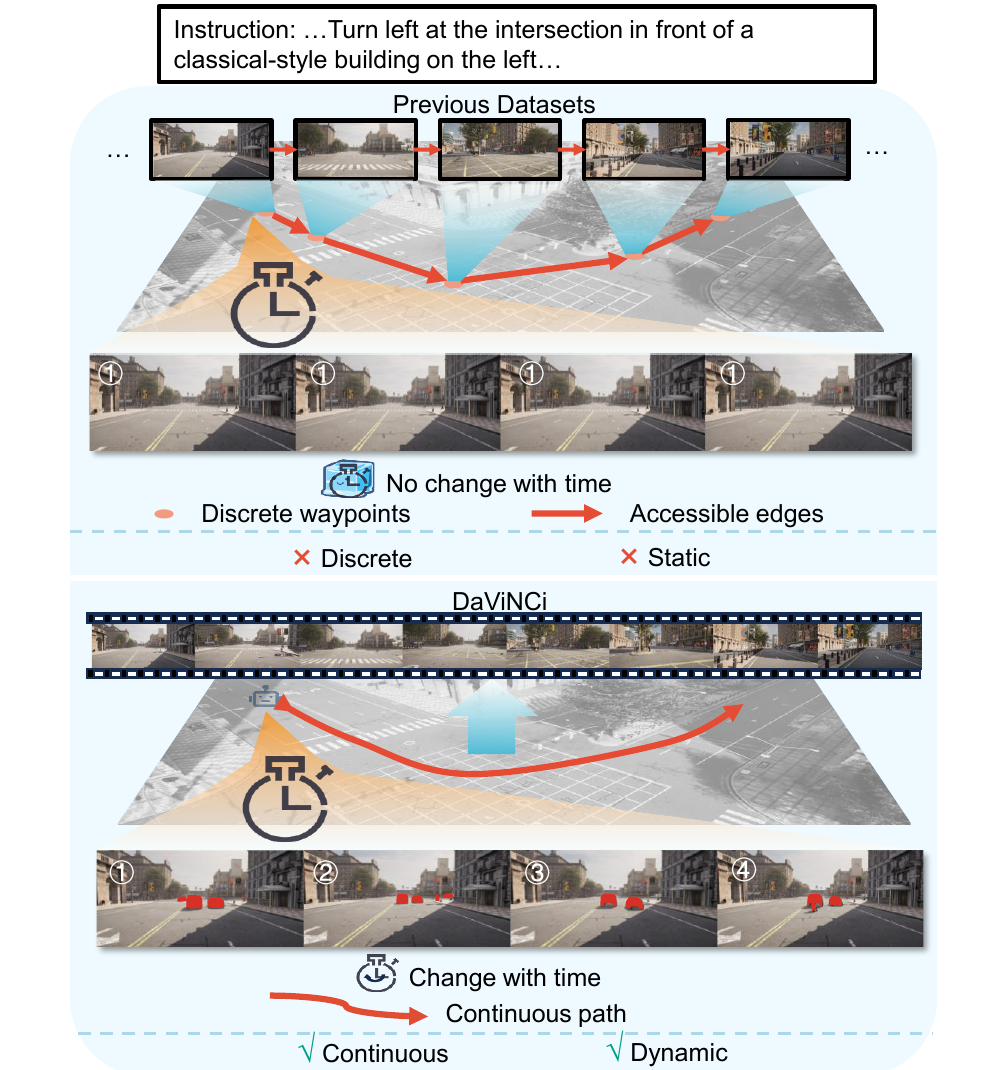}
  \caption{Comparison between DaViNCi and previous datasets. Unlike previous ones, it possesses continuous and dynamic properties}
  \label{fig:intro}
\end{figure}

\section{Introduction}
\begin{table*}[htbp]
\centering
\resizebox{\linewidth}{!}{%
\begin{tabular}{l c c c c c c c}
\toprule
Dataset & Environment & Trajectories & Instructions & Total length & Action space & Vocabulary & Dynamic\\
\midrule
R2R~\cite{anderson2018vision} & Indoor & 7,189 & 21,567 & 71.9K & Discrete & 3.1k & -- \\
R$×$R~\cite{ku2020room} & Indoor & 13,992 & 13,992 & 0.2M & Discrete & 7.0k & -- \\
CVDN~\cite{thomason2020vision} & Indoor & 7,415 & 2,050 & -- & Discrete & 4.4k & -- \\
REVERIE~\cite{qi2020reverie} & Indoor & 7,234 & 21,702 & 72.3K & Discrete & 1.6k & -- \\
SOON~\cite{zhu2021soon} & Indoor & 7,234 & 21,702 & 72.3K & Discrete & 1.6k & -- \\
VLN-CE~\cite{krantz2020beyond} & Indoor & 4,475 & 13,425 & 49.7K & Continuous & 4.3k & -- \\
\hline
TouchDown~\cite{chen2019touchdown} & Outdoor & 9,326 & 9,326 & 2.9M & Discrete & 5.0k & -- \\
Talk2Nav~\cite{vasudevan2021talk2nav} & Outdoor & 10,714 & 10,714 & -- & Discrete & 5.2k & -- \\
DaViNCi (Ours) & Outdoor & 6,933 & 6,933 & 2.6M & Continuous & 1.9k & $\sqrt{}$\\
\bottomrule
\end{tabular}%
}
\caption{Summary of the characteristic comparison between DaViNCi and the previous VLN datasets. "Dynamic" refers to the presence ($\sqrt{}$) or absence (--) of variable and interactive elements.}
\label{tab:datasets}
\end{table*}
When transporting passengers to an unfamiliar destination, one should inevitably follow verbal instructions to locate the correct route step by step. It would be convenient if vehicles could comprehend such instructions and operate autonomously. However, enabling an intelligent agent to execute such multimodal tasks in the complex outdoor environment remains a formidable challenge. The issue is caused by the relevant dataset for outdoor scenes still limited.

This navigation task is formally defined as Vision-and-Language Navigation (VLN) ~\cite{anderson2018vision}, which requires the agent to comprehend natural language instructions, involve visual observations, and navigate to the destination. To advance agents' capability in accomplishing this task, numerous datasets have been proposed to construct extensive VLN task scenarios. And a plenty of methods ~\cite{wang2024vision,chen2021history,chen2022think,an2022bevbert,wang2023gridmm,he2025mem4nav,xu2025flame,schumann2022analyzing,schumann2024velma} has been developed to address these datasets. Thus VLN has achieved considerable progress.

However, there remains an absence of datasets capable of accurately representing complex outdoor environments for vehicles. On one hand, the majority of existing VLN datasets ~\cite{anderson2018vision,ku2020room,qi2020reverie,zhu2021soon,krantz2020beyond} are confined to indoor settings, where visual and linguistic elements are sparser compared to outdoor environments, precluding straightforward adaptation to outdoor VLN tasks. On the other hand, even though there are widely used outdoor VLN datasets ~\cite{chen2019touchdown,schumann2021generating}, they generally suffer from two major limitations in representing outdoor environments as illustrated in Figure. \ref{fig:intro}: (1) \textbf{Fixed navigation waypoints}: Most outdoor VLN datasets employ fixed discrete topological graphs to constrain the agent's navigational space. This assumption disregards the characteristics of real-world vehicle mobility and obscures specific traffic regulations; (2) \textbf{Absence of dynamic elements}: The data is fixed alongside the topological structure, meaning that visual information at any location remains invariant over time. This is not compatible with the rapidly changing outdoor environment. These issues render existing VLN methods impractical for deployment in real-world outdoor vehicles.

To address this gap, we propose the DaViNCi (\textbf{D}yn\textbf{a}mic \textbf{Vi}sion-and-Language \textbf{N}avigation in \textbf{C}ont\textbf{i}nuous Environment) dataset. Rather than constructing paths through movement between fixed points, we directly delineate continuous vehicle trajectories in open maps. Furthermore, we capture visual data in real-time during the agent's operation to obtain dynamic information, rather than relying on pre-collected static visual content. Specifically, we utilize the CARLA simulator~\cite{Dosovitskiy17} to generate data. The trajectories of autonomous vehicles are tracked to constitute ground-truth paths, and corresponding natural language instructions are generated based on visual observations with the help of LLM. The dataset encompasses six distinct maps, comprising a total of 6,933 high-quality paths. To the best of our knowledge, DaViNCi represents the first outdoor VLN dataset incorporating dynamic elements. A summary and comparison of dataset characteristics are presented in Table. \ref{tab:datasets}. 

\begin{figure*}[t]
  \centering
  \includegraphics[width=\textwidth]{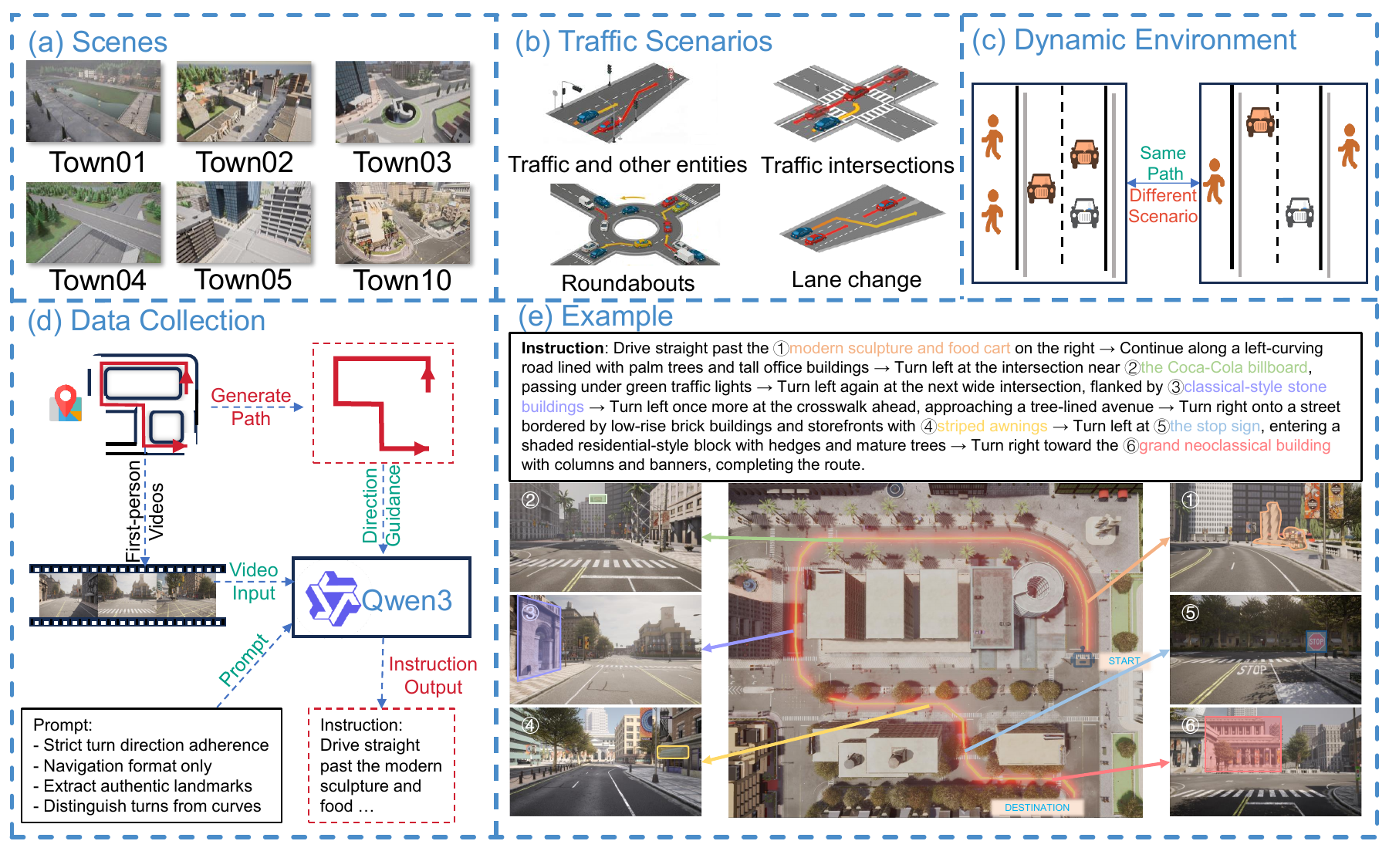}
  \caption{Overview of DaViNCi. (a) Several representative scenarios employed in the dataset. (b) Traffic scenarios that the agent vehicle may encounter during navigation . (c) The dynamic characteristics of the experimental environment . (d) The methodology for data collection and annotation. (e) A representative example demonstrating the alignment between textual and visual information.}
  \label{fig:dataset}
\end{figure*}

To facilitate experiments, we construct a novel continuous experimental environment, where the agent controls the vehicle through continuous actions based on real-time waypoints. During experiments, we introduce randomly generated autonomous vehicles and pedestrians as dynamic elements in the environment. We design a set of experiments to demonstrate the validity and challenges of DaViNCi. Initially, we discretize the dataset to align with the format of previous datasets, then compare the performance of the VLN methods on them. The success rate of approximately 30\%–40\% demonstrates the effectiveness of DaViNCi, while the over 10\% drop in success rate compared to previous datasets highlights its challenging nature. Subsequently, we switch to continuous environment. Considering that the majority of existing outdoor VLN methods cannot directly adapt to continuous environments, we introduce a reinforcement learning (RL)-based baseline approach. The lower success rate reflects the challenge posed by the continuous dynamic environment. Then we evaluate the impact of continuous action granularity and dynamic elements on task performance, revealing the reasons behind the challenge. In summary, our contributions are as follows:
\begin{itemize}
    \item We propose the DaViNCi (\textbf{D}yn\textbf{a}mic \textbf{Vi}sion-and-Language \textbf{N}avigation in \textbf{C}ont\textbf{i}nuous Environment) dataset, which achieves a paradigm shift from discrete to continuous and from static to dynamic representations in outdoor VLN datasets for vehicle.
    \item We propose test benchmarks of our dataset for both discrete and continuous environments.
    \item Through a series of experiments, we demonstrate both the validity and challenge of the dataset and provide a baseline in continuous environment.
\end{itemize}

\section{Related Work}
\textbf{Indoor VLN datasets}. The VLN task plays a crucial role in driving advancements in multimodal research. Since the R2R~\cite{anderson2018vision} dataset first formalized the concept of VLN tasks, the field has witnessed continuous expansion through the introduction of new datasets. Early VLN datasets were predominantly confined to indoor environments, encompassing various task formulations including step-by-step instruction following~\cite{anderson2018vision,jain2019stay,krantz2020beyond}, described goal navigation~\cite{qi2020reverie,zhu2021soon}, demand-driven navigation~\cite{wang2023find}, interactive navigation~\cite{shridhar2020alfred,yenamandra2023homerobot,li2023behavior}, and dialog-based navigation~\cite{thomason2020vision,gao2022dialfred}, thereby covering the majority of VLN application scenarios.  

\noindent\textbf{Outdoor VLN datasets}. Subsequent developments saw the extension of VLN datasets to outdoor environments. Notable examples include the Touchdown~\cite{chen2019touchdown} and Talk2Nav~\cite{vasudevan2021talk2nav} datasets, which were constructed using Google Street View imagery. Some datasets further diversified the experimental agents beyond ground vehicles, such as UAV~\cite{lee2025citynav,liu2023aerialvln} into the navigation paradigm.  
However, a persistent limitation remains in current vehicle-based outdoor VLN research: while indoor VLN has largely transitioned to continuous environments, outdoor navigation predominantly relies on discrete experimental settings. Moreover, existing outdoor datasets maintain completely static environments devoid of dynamic elements. To overcome these constraints, we introduces a novel continuous and dynamic outdoor VLN dataset, thereby expanding the task boundaries of VLN frameworks.

\section{Dataset}
We present an overview of the dataset in the figure. \ref{fig:dataset}.

\subsection{Construction Method}
\subsubsection{Scenario Selection}
Our scenario construction is built upon CARLA simulator. We selected six distinct outdoor maps $\{town01, \\ town02, town03, town04, town05, town10\}$ from the simulator to compose the dataset. The resulting collection contains 6,933 cases, covering a wide range of outdoor driving scenarios. 
$Town01$ and $town02$ are rural areas, with simple content but narrow horizons, fewer characteristic elements. Among them, $town02$ features a more complex network of roads. $Town03$ is a simple town structure, featuring some high buildings and unique road structures such as roundabouts, uphill and downhill. $Town04$ mainly includes a highway structure, which is long but monotonous. $Town05$ and $town10$ are urban areas possessing a complex road structure. They incorporate diverse landmark content such as buildings, traffic signs, billboards.
Furthermore, with reference to the CARLA AD Leaderboard ~\cite{Dauner2024ARXIV}, we designed following traffic scenarios during navigation: (1) Handling traffic lights and traffic signs and interacting with other vehicles, pedestrians, cyclists. (2) Negotiation at signalized and non signalized intersections. (3) Negotiation at roundabouts. (4) Lane changing to achieve correct path. The rich visual features support image based localization for the agent, while the complex and diverse structures challenge the agent’s path planning and memory capabilities. These characteristics guarantee the validity and diversity of the dataset.


\subsubsection{Data Collection and Annotation}
\textbf{Path Generation}. We obtain paths by tracking autonomous vehicles in CARLA. Specifically, in environments devoid of other dynamic elements, we initialize a random autonomous vehicle to navigate freely in the simulated world. During navigation, vehicle maneuvers at critical positions are determined by the autonomous driving program. Meanwhile, we collect the vehicle's position information at one-second intervals and then connect these waypoints to generate the ground-truth path for each traversal. Extensive path sampling contributes to comprehensive coverage of map elements. The resulting path distribution is as follows: $\{town01:969, town02:765, town03:930, town04:463, town05:951, town10:2855\}$.

\noindent\textbf{Instruction Generation and Alignment}. We employ a Visual Language Model (VLM) to generate instructional content and then perform manual fine-tuning. During the vehicle operation process described above, first-person driving videos are recorded to provide source material for instruction generation. To address the limitations of existing VLMs in generating navigation descriptions from first-person driving videos, this paper proposes a two-stage processing framework. We first accurately extracts turn events based on trajectory data as prior constraints, then integrates multi-modal visual analysis to generate structured navigation instructions. Ultimately, a natural language navigation sequence aligned with driving logic and containing landmark details is produced with the help of manual fine-tuning. This method effectively enhances the accuracy and practicality of navigation descriptions through a synergistic "mathematical quantification + visual semantic generation" mechanism.

\noindent\textbf{Visual Information Acquisition}. We collect visual information in real-time. Previous outdoor VLN data-sets typically require a discrete topological graph to define the reachable waypoints. Each waypoint is accompanied by a panoramic image to provide visual information. In contrast, we emphasizes dynamic environments, where visual information acquired at the same location may vary. During navigation, whenever the agent requests visual information, the system provides first-person perspective images in real-time, thereby fulfilling the dynamic requirements.

\subsubsection{Dynamic Elements of Environment}
With the path-instruction pairs fixed, we dynamically adjust the elements in the simulated world. Specifically, at the beginning of each experimental trial, we introduce other vehicles and pedestrians as dynamic elements in the environment. These obstacles act based on programmatic rules and operate randomly. The agent must process varying environmental information even when navigating identical paths.

\subsubsection{Quality Control}
We employed the same LLM and prompts for instruction generation, which ensures the uniformity of annotation. We also performed preliminary screening to identify navigation text descriptions that exhibited logical inconsistencies or unreasonable content. Furthermore, we eliminated unreasonable path instances caused by collisions and traffic violations. Ultimately, we obtained 6,933 high-quality path-instruction pairs.

\subsection{Data Statistics}
\begin{figure}[t]
  \centering
  \includegraphics[width=\linewidth]{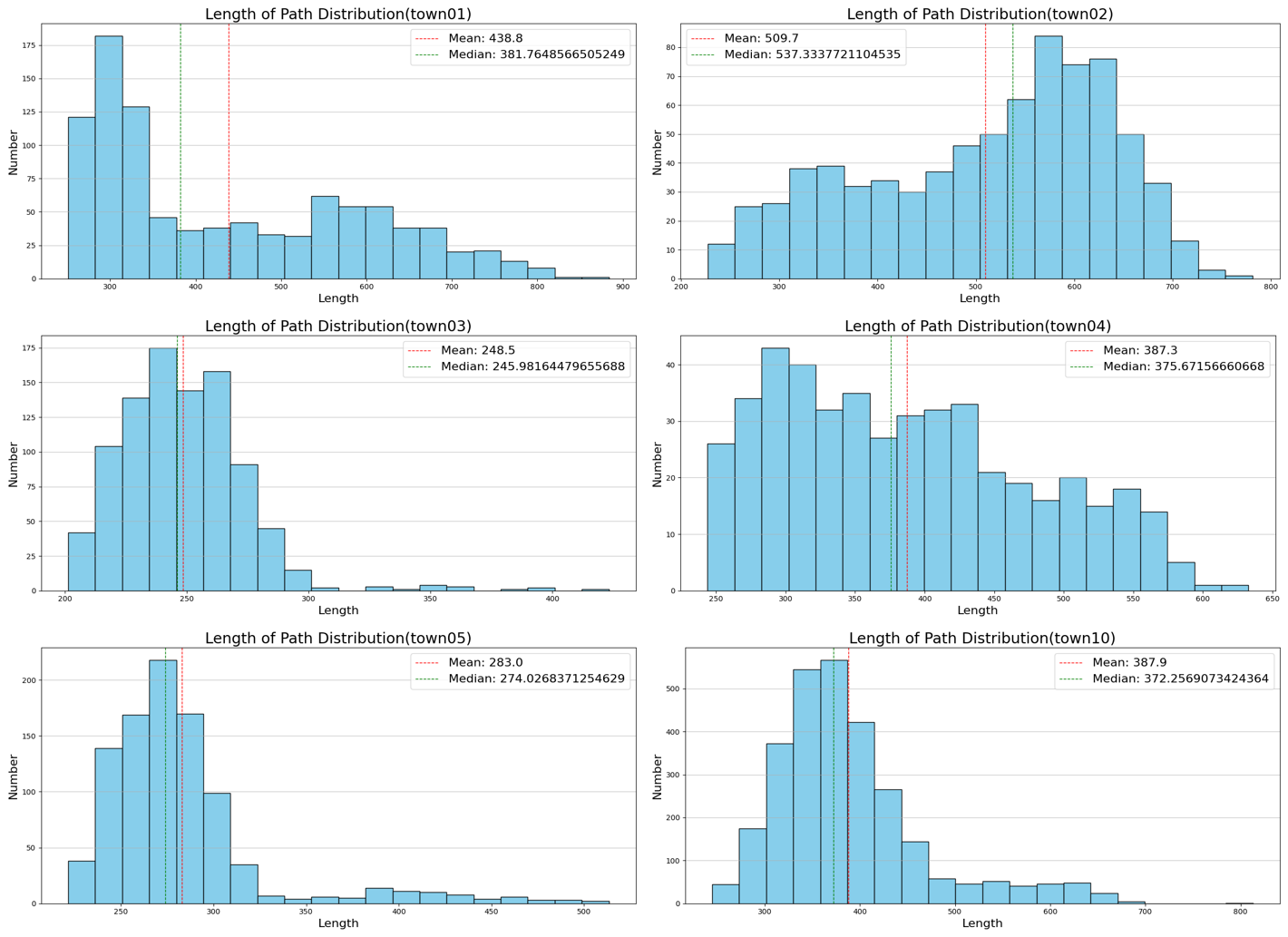}
  \caption{Statistics of Path Length. The distribution of path lengths is displayed across different maps. The red line represents the average, while the green line represents the median.}
  \label{fig:Statistics of Path Length}
\end{figure}

\textbf{Path Length}. The figure. \ref{fig:Statistics of Path Length} illustrates the distribution of path lengths (in meters) across the six maps in DaViNCi. These maps encompass paths of varying lengths, ranging from approximately 200 m to 800 m. According to the characteristics of each map, path lengths are generally concentrated around specific values. The distribution enables the dataset to simulate navigation routes of different difficulty levels and types. Generally, longer-distance navigation poses greater challenges. 

\begin{figure}[t]
  \centering
  \includegraphics[width=\linewidth]{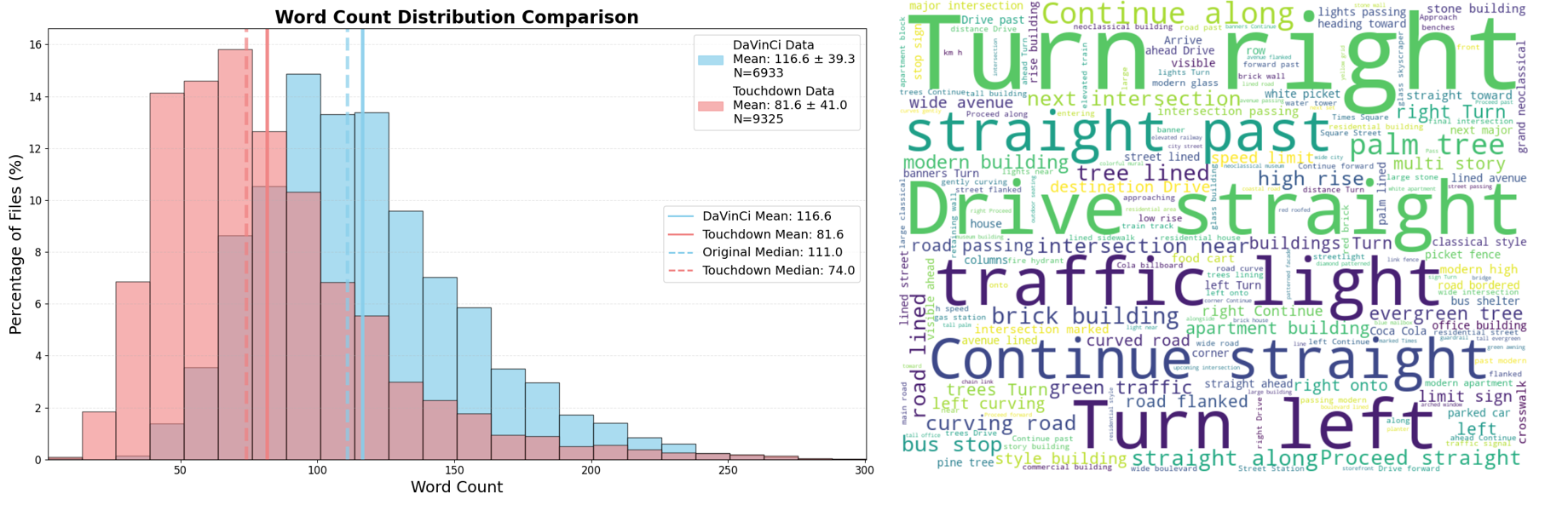}
  \caption{Statistics of Instructions. The left subfigure illustrates the comparison of text length distributions between our dataset and the Touchdown dataset. The right subfigure displays the frequency of high-frequency words in our dataset, where larger font size indicates higher occurrence frequency.}
  \label{fig:Statistics of Instructions}
\end{figure}

\noindent\textbf{Text Length}. The figure. \ref{fig:Statistics of Instructions} presents the statistical information of instructions. The text length in our dataset is generally longer than that in Touchdown, posing greater challenges for text-vision alignment in the agent. In our dataset, directional instruction terms (e.g., "turn left," "turn right," "straight") appear with high frequency, indicating that the instructions actively contribute to action guidance. Additionally, several landmark-related terms are present, demonstrating how textual descriptions guide the agent in localization.

\begin{table*}[t]
  \centering
    \resizebox{\linewidth}{!}{%
    \begin{tabular}{lcccccccccccc|ccccccccc}
      \toprule
      \multirow{2}{*}{Model} & \multicolumn{3}{c}{town01} & \multicolumn{3}{c}{town02} & \multicolumn{3}{c}{town03} & \multicolumn{3}{c}{town10} & \multicolumn{3}{c}{Touchdown} & \multicolumn{3}{c}{Map2Seq} \\
      \cmidrule(lr){2-4} \cmidrule(lr){5-7} \cmidrule(lr){8-10} \cmidrule(lr){11-13} \cmidrule(lr){14-16} \cmidrule(lr){17-19}
      & TC$\uparrow$ & SPD$\downarrow$ & nDTW$\uparrow$ & TC$\uparrow$ & SPD$\downarrow$ & nDTW$\uparrow$ & TC$\uparrow$ & SPD$\downarrow$ & nDTW$\uparrow$ & TC$\uparrow$ & SPD$\downarrow$ & nDTW$\uparrow$ & TC$\uparrow$ & SPD$\downarrow$ & nDTW$\uparrow$ & TC$\uparrow$ & SPD$\downarrow$ & nDTW$\uparrow$ \\
      \midrule
      GA (2019)          & 1.0  & 61.53 & 0.42  & 6.6  & 26.89 & -     & 8.8  & 9.11  & 32.51 & 8.1  & 28.07 & -     & 11.9 & 19.00 & 24.90 & \textbf{17.0} & - & 30.10 \\
      Rconcat (2019)     & 1.0  & 61.43 & 0.72  & 6.6  & 28.57 & -     & 6.6  & 8.64  & 37.60 & 8.5  & 29.71 & -     & 11.8 & 20.40 & 22.90 & \textbf{14.7} & - & 27.70 \\
      VLN-trans (2021)   & 15.6 & 45.08 & -     & 7.9  & 25.87 & 2.35  & 9.9  & 10.07 & 30.15 & 6.3  & 29.54 & -     & 16.2 & 20.80 & 27.80 & \textbf{17.0} & - & 29.50 \\
      ORAR-pre-final (2022)   & 10.4 & 39.49 & 10.98 & 14.5 & 24.68 & -     & 7.7  & 10.69 & 29.59 & 15.5 & 19.39 & 9.88  & - & - & - & - & - & -\\
      ORAR-4th-to-last (2022)    & 9.4  & 40.58 & 10.09 & 3.9  & 28.59 & 0.08  & 14.3 & 7.79  & 45.69 & 15.5 & 20.89 & 9.89  & 29.6 & 11.79 & 45.30 & \textbf{47.8} & 6.53 & 62.10 \\
      VELMA (2024)       & 17.7 & 16.28 & 37.19 & 15.8 & 16.39 & 34.29 & 24.2 & 15.49 & 40.39 & 19.7 & 17.99 & 38.79 & 27.4 & 15.03 & 41.93 & \textbf{48.7} & 6.80 & 62.37 \\
      VELMA+Mem4Nav (2025)& 21.9 & 13.89 & 46.89 & 19.7 & 14.19 & 43.99 & 28.6 & 13.29 & 51.79 & 24.3 & 15.59 & 49.29 & 34.0 & 12.90 & 48.82 & \textbf{56.8} & 6.10 & 72.71 \\
      FLAME (2025)       & 26.0 & 9.98  & 50.39 & 23.7 & 10.29 & 48.69 & 34.1 & 9.49  & 59.89 & 29.9 & 10.79 & 52.19 & 40.2 & 9.53  & 54.56 & \textbf{52.4} & 5.91 & 67.72 \\
      FLAME+Mem4Nav (2025)& 31.3 & 9.39 & 60.19 & 28.9 & 9.59 & 58.29 & 40.7 & 8.79 & 70.29 & 35.9 & 9.99 & 61.49 & 48.5 & 9.10 & 63.63 & \textbf{60.4} & 5.90 & 75.94 \\
      \bottomrule
    \end{tabular}
  }
  \caption{Performance of Models on datasets (Test Set)}
  \label{tab:hard_scene_town_test}
\end{table*}
\begin{table}[t]
  \centering
  \resizebox{\linewidth}{!}{
  \begin{tabular}{lccccccccc}
    \toprule
    \multirow{2}{*}{Methods} & \multicolumn{3}{c}{town01} & \multicolumn{3}{c}{town02} & \multicolumn{3}{c}{town03} \\
    \cmidrule(lr){2-4} \cmidrule(lr){5-7} \cmidrule(lr){8-10}
    & ANE$\downarrow$ & SR$\uparrow$ & APC$\uparrow$ & ANE$\downarrow$ & SR$\uparrow$ & APC$\uparrow$ & ANE$\downarrow$ & SR$\uparrow$ & APC$\uparrow$ \\
    \midrule
    Our Method & 79.86 & 13.2 & 40.4 & 92.10 & 5.8 & 28.0 & 102.51 & 17.6 & 39.7 \\
    Random (-stop) & 134.81 & 3.5 & 18.6 & 102.43 & 2.1 & 13.5 & 145.37 & 3.4 & 20.3 \\
    Random & 204.97 & 0.0 & 2.1 & 119.80 & 0.0 & 1.1 & 161.40 & 0.0 & 1.8 \\
    \midrule
    & \multicolumn{3}{c}{town04} & \multicolumn{3}{c}{town05} & \multicolumn{3}{c}{town10} \\
    \cmidrule(lr){2-4} \cmidrule(lr){5-7} \cmidrule(lr){8-10}
    & ANE$\downarrow$ & SR$\uparrow$ & APC$\uparrow$ & ANE$\downarrow$ & SR$\uparrow$ & APC$\uparrow$ & ANE$\downarrow$ & SR$\uparrow$ & APC$\uparrow$ \\
    \midrule
    Our Method & 102.01 & 12.7 & 44.3 & 131.87 & 12.0 & 33.4 & 100.16 & \textbf{18.2} & \textbf{48.0} \\
    Random (-stop) & 186.28 & 3.2 & 15.8 & 158.24 & 2.8 & 16.9 & 117.36 & 2.1 & 23.7 \\
    Random & 292.54 & 0.0 & 2.4 & 169.60 & 0.0 & 2.7 & 129.60 & 0.3 & 1.3 \\
    \bottomrule
  \end{tabular}
  }
  \caption{Comparison of Our Method and Random in DaViNCi. "-stop" means no $stop$ in action space.}
  \label{tab:town_compare}
\end{table}
\begin{table}[t]
  \centering
  \resizebox{\linewidth}{!}{
  \begin{tabular}{cccccccccc}
    \toprule
    \multirow{2}{*}{Granularity} & \multicolumn{3}{c}{town01} & \multicolumn{3}{c}{town03} & \multicolumn{3}{c}{town10}\\
    \cmidrule(lr){2-4} \cmidrule(lr){5-7} \cmidrule(lr){8-10}
    & ANE$\downarrow$ & SR$\uparrow$ & APC$\uparrow$ & ANE$\downarrow$ & SR$\uparrow$ & APC$\uparrow$ & ANE$\downarrow$ & SR$\uparrow$ & APC$\uparrow$ \\
    \midrule
    3m  & 77.62 & 13.6 & 42.1 & 99.81 & 17.1 & 40.3 & 101.45 & 18.9 & 47.3\\
    4m  & 79.86 & 13.2 & 40.4 & 102.51 & 17.6 & 39.7 & 100.16 & 18.2 & 48.0\\
    5m  & 78.04 & 12.9 & 39.7 & 103.27 & 17.9 & 41.8 & 98.64 & 17.5 & 50.1\\
    \bottomrule
  \end{tabular}
  }
  \caption{Performance under different granularities}
  \label{tab:granularity}
\end{table}
\begin{table}[t]
  \centering
  \resizebox{\linewidth}{!}{
  \begin{tabular}{cccccccccc}
    \toprule
    \multirow{2}{*}{Dynamic Elements} & \multicolumn{3}{c}{town01} & \multicolumn{3}{c}{town03} & \multicolumn{3}{c}{town10}\\
    \cmidrule(lr){2-4} \cmidrule(lr){5-7} \cmidrule(lr){8-10}
    & ANE$\downarrow$ & SR$\uparrow$ & APC$\uparrow$ & ANE$\downarrow$ & SR$\uparrow$ & APC$\uparrow$ & ANE$\downarrow$ & SR$\uparrow$ & APC$\uparrow$ \\
    \midrule
    None  & 72.11 & 17.9 & 48.6 & 90.25 & 21.5 & 49.6 & 87.36 & 23.1 & 57.8\\
    10 vehicles and 10 pedestrians & 73.51 & 15.3 & 44.1 & 97.46 & 19.2 & 43.3 & 94.17 & 21.7 & 51.3\\
    30 vehicles and 30 pedestrians & 79.86 & 13.2 & 40.4 & 102.51 & 17.6 & 39.7 & 100.16 & 18.1 & 48.0\\
    \bottomrule
  \end{tabular}
  }
  \caption{Performance under different dynamic numbers}
  \label{tab:dynamic_num}
\end{table}

\section{Experiment}
\subsection{Discrete environmental Experiments}
Prior to conducting continuous environment experiments, we discretize the dataset to evaluate the performance of previous outdoor VLN methods on DaViNCi.

\subsubsection{Discretization Adaptation} We select four maps from the dataset $(town01, town02, town03, and town10)$ to construct fixed discrete topological graph structures. The data is formatted following the Touchdown dataset. Notably, we apply different granularities across different maps. Specifically, the $town01$, $town02$, and $town10$ maps adopt a fine granularity of 5 meters, which captures more details of the environment and increases the number of actions required for navigation. In contrast, the $town03$ map employs a coarse granularity of 12 meters, which is similar to that used in previous outdoor VLN datasets. Finally, we adapt continuous driving paths into ordered discrete point sequences to obtain the agent's discrete trajectories.

\subsubsection{Experiment Setup}
\textbf{Dataset Splitting}. We randomly partition the trajectories from the four maps into training, validation, and test sets following an approximately 8:1:1 ratio, respectively. They possess similar distributions of trajectory lengths and instruction complexities.

\noindent\textbf{Evaluation Metrics}. We adopt three standard evaluation metrics from prior work ~\cite{xu2025flame, he2025mem4nav}: TC, SPD, nDTW.

\noindent\textbf{Implementation details}. All models are trained using the optimizer with learning rate $\eta = 1.5×10^{-3}$, batch size $B=16$, and distributed across four NVIDIA RTX 3090 GPUs.

\subsubsection{Results}
The experimental results on the test set are presented in Table. \ref{tab:hard_scene_town_test}, along with the performance of these models on the Touchdown and Map2Seq~\cite{schumann2021generating} test sets for comparison. The results indicate that the models generally exhibit lower performance on our dataset compared to previous datasets, which demonstrates that our dataset introduces new challenges. Nevertheless, the overall experimental performance on our dataset remains considerable, thereby validating the effectiveness of the dataset.

When it comes to internal comparison, we observe that town03, which employs coarse granularity, achieves superior experimental results. This can be attributed to reduced complexity of information. The performance on the other three maps has slightly decreased, and their TC variation trends show a certain positive correlation with path length. This also reflects the challenging nature of long-distance navigation for VLN tasks. 

\subsection{Continuous Environmental Experiment}
We developed a RL-based framework that enables agents to dynamically respond to environmental changes during navigation. The approach builds upon established RL paradigms ~\cite{zhang2025activevln} while introducing specific refinements to accommodate continuous dynamic environment.

\subsubsection{Data Composition}
The VLN requires the agent to navigate in the CARLA simulation environment by outputting actions $\in \mathcal{A}$ at each step. Visual$\mathcal{V} $  and textual inputs $\mathcal{I}$ would be provided. The agent ultimately reach the specified destination along a predetermined path.

\noindent\textbf{Action Space}: 
The action space is defined as: 
\[
\mathcal{A}\in\{sharpleft, left, straight, right, sharpright, lanechange, stop\}
\]

In contrast to previous VLN datasets, where turning actions typically involve in-place rotations, our formulation directly advances the agent toward the specified direction. During operation, the agent’s available actions are dynamically constrained by real-time road condition monitoring, ensuring compliance with traffic rules.

\noindent\textbf{Visual Input}: The agent's state is represented by spatial coordinates $\mathbf{p} = (x, y, z) \in \mathbb{R}^3$ , supplemented by orientation angles $(\theta, \phi) \in [0, 360^\circ)^2$ denoting yaw and pitch, respectively. These parameters enable the environment to determine the agent’s required visual perception. At each timestep $t$, the simulator captures a first-person RGB image $\mathcal{V}_t$ (1280×720 resolution) based on the agent's current state $\mathcal{S}_t = (x, y, z, \theta, \phi)$.

\noindent\textbf{Textual Guidance}: Each path is associated with a natural language navigation instruction $\mathcal{I} = \{w_1, w_2, ..., w_n\}$, providing step-by-step directional and landmark references.

\begin{figure}[t]
  \centering
  \includegraphics[width=\linewidth]{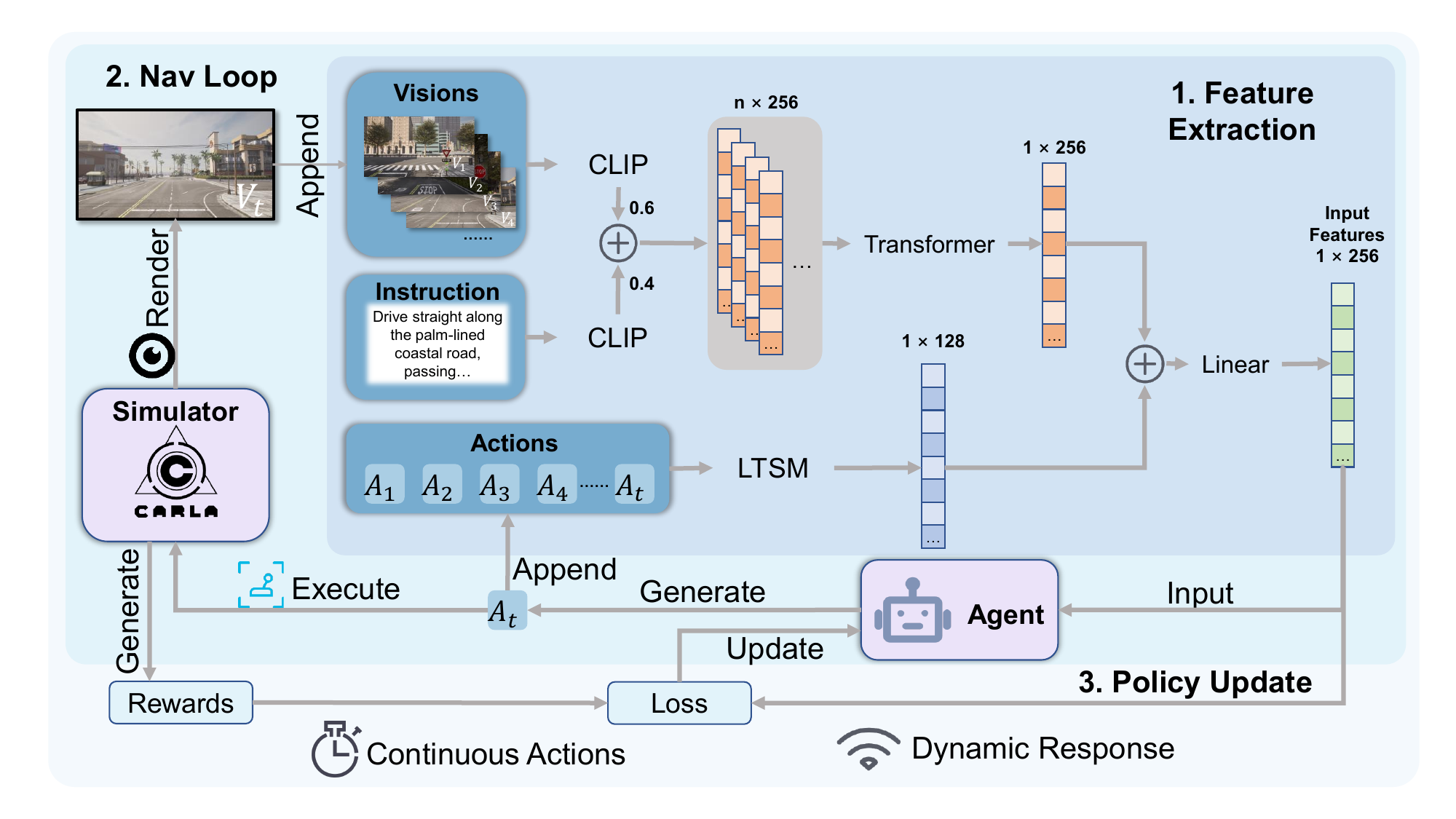}
  \caption{COVL-RL Architecture. (1) Feature extraction and content integration of input content. (2) Navigation loop for each path. (3) Policy update process of the agent.}
  \label{fig:method}
\end{figure}

\subsubsection{COVL-RL}
 In the DaViNCi setting, visual observations are generated online during navigation, requiring the agent to learn continuously from interactive experience. To this end, we propose \textbf{COVL-RL}, a context-aware online vision-language reinforcement learning framework. It supports dynamic response and policy learning from self-generated trajectory experience in continuous environment. The Architecture is shown in figure. \ref{fig:method}. At each time step $t$, the agent takes as input the global textual instruction, current visual observation, and historical trajectory context to infer the next action:
\begin{equation}
a_{t}\sim\pi_{\theta}\left(a_{t}\mid\mathcal{I},\mathcal{H}_{t},\mathcal{V}_{t}\right)
\end{equation}

where $\mathcal{I}$ denotes the natural language navigation instruction, $\mathcal{V}_{t}$ is the visual observation at step $t$, and $\mathcal{H}_{t}$ represents the historical trajectory context up to the previous step.
Specifically, the context $\mathcal{H}_{t}$ integrates past visual inputs and executed actions:
\begin{equation}
\mathcal{H}_{t}=\{\mathcal{V}_{1},A_{1},\mathcal{V}_{2},A_{2},\ldots,\mathcal{V}_{t-1},A_{t-1}\},
\end{equation}

in which $\mathcal{V}_t$ and $A_t$ denote the visual observation and action at time $t$, respectively.

\subsubsection{Experiment Setup}
\textbf{Dataset Splitting}. We designate only the town10 map as the training environment, randomly partitioning it into training, validation, and test sets following a ratio of approximately 6:1:1. The remaining five maps are all treated as unseen scenes for testing.

\noindent\textbf{Evaluation Metrics}. We adapt and extend the evaluation metrics from the discrete environment, resulting in the following new metrics for the continuous environment: 
\textbf{Average Navigation Error(m)}: Mean Euclidean distance between the agent's final position and the target.
\textbf{Success Rate(\%)}: The percentage of instances where the final position falls within a 10-meter radius of the end point.
\textbf{Average Path Completion(\%)}: Average proportion of the reference path traversed by the agent.
\begin{equation}
\text{Path Completion (PC)} = \frac{L_{\text{traversed}}}{L_{\text{total}}}
\end{equation} 

\noindent\textbf{Implementation details}. We use a learning rate $\eta = 3×10^{-4}$ for the method. The action is executed with step size 4 meters per timestep, and the environment contains 30 dynamic pedestrians and 30 vehicles as dynamic elements by default.

\subsubsection{Results}
The primary experimental results are presented in Table. \ref{tab:town_compare}, where we compare our proposed baseline method with the random approach. The first random method refers that the agent selects actions with equal probability from the action space. The second random method eliminates "stop" from the action space and uses a program to halt by detecting deviation from the route. As shown in the table, the random method achieves virtually no successful cases, demonstrating extremely poor path completion performance. In contrast, our baseline method can complete tasks on certain trajectories, achieving a considerable path completion rate. Performance on unseen environments is generally inferior to that on seen environments, demonstrating the challenge this data partition method poses to the model's generalization capability. The success rate here is lower compared to that in discrete environment experiments, reflecting the challenges posed by the continuous dynamic environment setting for VLN tasks.

\noindent\textbf{Impact of Granularity}. We conduct experiments with the agent executing tasks at different granularities, specifically 3m, 4m, and 5m per action. The experimental results are presented in Table. \ref{tab:granularity}. It can be observed that the variation in granularity does not lead to significant changes in experimental performance. This can be explained that finer granularity enables the agent to receive more environmental information, whereas coarser granularity reduces the number of action steps. The trade-off results in granularity having a non-linear impact on task performance.

\noindent\textbf{Impact of Dynamic Elements}. We instruct the agent to perform tasks with varying quantities of dynamic elements. Specifically, three different dynamic environments are configured: (1) No dynamic elements, (2) 10 vehicles and 10 pedestrians, and (3) 30 vehicles and 30 pedestrians. The results are presented in Table. \ref{tab:dynamic_num}. As the number of dynamic elements increases, the agent's task completion performance progressively declines. This indicates that these dynamic elements genuinely affect the agent's capability to process environmental information, presenting new challenges for VLN tasks. Through case analysis, we find the challenge stems from the obstruction of dynamic elements to the partial field of view of the agent. It needs to adapt to such visual field defects.

\section{Conclusion}
In this paper, we propose the DaViNCi dataset, the first outdoor VLN dataset that simultaneously introduces both continuous and dynamic factors. Our dataset encompasses six distinct maps covering various scenarios with a total of 6,933 trajectories, and introduces a novel data collection approach to ensure both the effectiveness and diversity of the dataset. Furthermore, we construct a dynamic experimental environment using the CARLA simulator to conduct experiments. We design experiments for both discrete and continuous environments. The former validates existing outdoor VLN methods on DaViNCi and compares the performance with other datasets. A decrease in success rate of more than 10\% preliminarily demonstrating the challenge level of DaViNCi. The latter highlights the impact of continuous action granularity and dynamic elements on VLN tasks, which further illustrates the challenge level of the dataset. In summary, the introduction of DaViNCi will significantly advance the sim-to-real transfer of VLN agents.

\bibliographystyle{ACM-Reference-Format}
\bibliography{arefer}

\appendix

\end{document}